\documentclass[letterpaper, 10 pt, journal, twoside]{IEEEtran}

\IEEEoverridecommandlockouts                              
\usepackage{times}
\usepackage{amsthm}
\usepackage{amsmath,amsfonts,amssymb}
\usepackage{prettyref}
\usepackage{mathrsfs}
\usepackage{graphicx}
\usepackage{wrapfig}
\usepackage{subfig}
\usepackage{MnSymbol}
\usepackage{multirow}
\usepackage{booktabs}
\usepackage{algorithm}
\usepackage[noend]{algpseudocode}
\usepackage[table]{xcolor}
\usepackage{cellspace}
\usepackage
[backend=bibtex,
bibstyle=ieee,
citestyle=numeric,
mincitenames=1,
maxcitenames=2,
natbib=true,
doi=false,
isbn=false,
url=false,
eprint=false]{biblatex}
\usepackage[colorlinks,allcolors=gray,hypertexnames=true]{hyperref} 
\newcommand{\citewithauthor}[1]{\citeauthor{#1} \cite{#1}}
\usepackage{diagbox}

\usepackage{pgf}
\pgfkeys{/pgf/number format/.cd ,precision=2,sci generic={exponent={\times 10^{#1}}}}

\newtheorem{theorem}{Theorem}[section]

\newtheorem{corollary}[theorem]{\TE{Corollary}}

\algnewcommand{\LineComment}[1]{\State \(\triangleright\) #1}
\algdef{SE}[DOWHILE]{Do}{doWhile}{\algorithmicdo}[1]{\algorithmicwhile\ #1}
\newcommand*{\colorboxed}{}
\def\colorboxed#1#{%
  \colorboxedAux{#1}%
}
\newcommand*{\colorboxedAux}[3]{%
  \begingroup
    \colorlet{cb@saved}{.}%
    \color#1{#2}%
    \boxed{%
      \color{cb@saved}%
      #3%
    }%
  \endgroup
}

\newrefformat{fig}{Figure~\ref{#1}}
\newrefformat{par}{Section~\ref{#1}}
\newrefformat{appen}{Appendix~\ref{#1}}
\newrefformat{sec}{Section~\ref{#1}}
\newrefformat{sub}{Section~\ref{#1}}
\newrefformat{table}{Table~\ref{#1}}
\newrefformat{ass}{Assumption~\ref{#1}}
\newrefformat{alg}{Algorithm~\ref{#1}}
\newrefformat{def}{Definition~\ref{#1}}
\newrefformat{cor}{Corollary~\ref{#1}}
\newrefformat{thm}{Theorem~\ref{#1}}
\newrefformat{lem}{Lemma~\ref{#1}}
\newrefformat{step}{Step~\ref{#1}}
\newrefformat{ln}{Line~\ref{#1}}
\newrefformat{eq}{Equation~\ref{#1}}
\newrefformat{pb}{Problem~\ref{#1}}
\newrefformat{it}{Item~\ref{#1}}
\newrefformat{te}{Term~\ref{#1}}
\def\Eqref Eq:#1:{\eqref{eq:#1}}
\newrefformat{Eq}{Equation~\Eqref#1:}

\newcommand{\TE}[1]{\textbf{#1}}

\newcommand{\FPP}[2]{\frac{\partial{#1}}{\partial{#2}}}

\newcommand{\FPPT}[2]{\frac{\partial^2{#1}}{\partial{#2}^2}}

\newcommand{\FPPTT}[3]{\frac{\partial^2{#1}}{\partial{#2}\partial{#3}}}

\newcommand{\FDD}[2]{\frac{d{#1}}{d{#2}}}
\newcommand{\FDDT}[2]{\frac{d^2{#1}}{d{#2}^2}}

\newcommand{\TWO}[2]{\left(\setlength{\arraycolsep}{1pt}\begin{array}{cc}{#1}, & {#2}\end{array}\right)}
\newcommand{\TWOC}[2]{\left(\setlength{\arraycolsep}{1pt}\begin{array}{c}#1 \\ #2\end{array}\right)}
\newcommand{\TWOR}[2]{\left(\setlength{\arraycolsep}{1pt}\begin{array}{cc}{#1}^T, & {#2}^T\end{array}\right)^T}
\newcommand{\THREE}[3]{\left(\setlength{\arraycolsep}{1pt}\begin{array}{ccc}{#1}, & {#2}, & {#3}\end{array}\right)}

\newcommand{\FOUR}[4]{\left(\setlength{\arraycolsep}{1pt}\begin{array}{cccc}{#1}, & {#2}, & {#3}, & {#4}\end{array}\right)}

\newcommand{\MTT}[4]{\left(\setlength{\arraycolsep}{1pt}\begin{array}{cc}#1 & #2 \\ #3 & #4\end{array}\right)}

\newcommand{\dist}{\text{dist}}

\newcommand{\argmin}[1]{\underset{#1}{\text{argmin}}\;}
\newcommand{\argminP}[1]{\text{argmin}\;}

\newcommand{\argmaxP}[1]{\text{argmax}\;}
\newcommand{\ST}{\text{s.t.}\;}

\newcommand{\proofread}[1]{}
\newif\ifArxiv
\usepackage{xcolor}
\definecolor{Blue}{rgb}{0.2, 0.2, 0.8}
\definecolor{Black}{rgb}{0.0, 0.0, 0.0}

\makeatletter
\IEEEtriggercmd{\reset@font\normalfont\fontsize{8.8pt}{8.8pt}\selectfont}
\makeatother
\IEEEtriggeratref{1}

\title{\Large{Second-Order Convergent Collision-Constrained Optimization-Based Planner}}

\author{Chen Liang$^1$, Xifeng Gao$^2$, Kui Wu$^2$, Zherong Pan$^{2\dagger}$\vspace{-20px}\\
\thanks{\footnotesize{$^1$Chen Liang is with the department of computer science, Zhejiang university. $^2$ Xifeng Gao, Kui Wu, and Zherong Pan are with LightSpeed Studios, Tencent America. \{xifgao,kwwu,zrpan\}@global.tencent.com. $^\dagger$ indicates corresponding author.}}}
\newcommand\numberthis{\addtocounter{equation}{1}\tag{\theequation}}
\IEEEaftertitletext{\vspace{-1.\baselineskip}}
\allowdisplaybreaks
\begin{document}
\maketitle

\newif\ifdetail
\detailfalse

\begin{abstract}
Finding robot poses and trajectories represents a foundational aspect of robot motion planning. Despite decades of research, efficiently and robustly addressing these challenges is still difficult. Existing approaches are often plagued by various limitations, such as intricate geometric approximations, violations of collision constraints, or slow first-order convergence. In this paper, we introduce two novel optimization formulations that offer provable robustness, achieving second-order convergence while requiring only a convex approximation of the robot's links and obstacles. Our first method, known as the Explicit Collision Barrier (ECB) method, employs a barrier function to guarantee separation between convex objects. ECB uses an efficient matrix factorization technique, enabling a second-order Newton's method with an iterative complexity linear in the number of separating planes. Our second method, referred to as the Implicit Collision Barrier (ICB) method, further transforms the separating planes into implicit functions of robot poses. We show such an implicit objective function is twice-differentiable, with derivatives evaluated at a linear complexity. To assess the effectiveness of our approaches, we conduct a comparative study with a first-order baseline algorithm across six testing scenarios. Our results unequivocally justify that our method exhibits significantly faster convergence rates compared to the baseline algorithm.
\end{abstract}

\section{\label{sec:introduction}Introduction}
The issue of robot safety stands as a paramount concern, imposing stringent constraints on motion planning algorithms. Among these constraints, collision avoidance requirements demand that the robot maintains a minimum distance from static or moving obstacles at all times. This paper addresses the challenge of generating optimal robot poses and trajectories while adhering to collision constraints and optimizing a user-defined cost function. Such a problem finds practical applications in (multi-)UAV trajectory planning~\cite{deits2015efficient}, inverse kinematics~\cite{6628553}, object settling~\cite{hauser2021semi}, and dynamic simulations~\cite{khazoom2022humanoid}. Ideally, an effective planning algorithm should possess several key attributes to cater to the diverse needs of these applications: (Scalability) handling a significant number of robot links and obstacles; (Efficacy) rapid convergence and low iterative and overall computational costs; (Robustness) guarantee to satisfy all collision constraints; (Generality) applying to arbitrary articulated robots with complex geometries. However, despite decades of dedicated research in pursuit of such algorithms, it remains a challenge to find a single solution that fully satisfies all these diverse requirements.

Existing trajectory generation algorithms are engineered to suit a set of specific applications. Take, for instance, (online) UAV trajectory generation, where computational efficiency is paramount. Existing algorithm~\cite{gao2017gradient} employs penalty methods to handle constraints, which can compromise robustness. In contrast, other algorithms such as~\cite{deits2015efficient,liu2017planning} offer guaranteed robustness and global optimality. Unfortunately, their applicability is limited to point robots and convex-decomposable freespace. When dealing with general articulated robots, the formulation of collision constraints becomes notably challenging. Many optimization-based algorithms, as seen in the works~\cite{kalakrishnan2011stomp,zucker2013chomp}, once again resort to less robust penalty-based techniques. Furthermore, these algorithms often exhibit limited efficacy, having only first-order convergence. Recent advancements~\cite{li2020incremental,zhang2023provably} have introduced interior point methods to address collision constraints with guaranteed robustness. However, \citewithauthor{li2020incremental} necessitates the decomposition of robot geometry into a gazillion of triangles, which scales poorly with the complexity of the robot geometry. On the other hand, \citewithauthor{zhang2023provably} relies on the strict convexity of robot links, demanding intricate geometric modifications and suffering from first-order convergence limitations.

\begin{table}[ht]
\centering
\begin{tabular}{ccccc}
\toprule
Method & Scalability & Efficacy & Robustness & Generality\\
\midrule
\cite{gao2017gradient} & point-penalty & second-order & no & point \\
\cite{deits2015efficient,liu2017planning} & convex & mixed-integer & yes & point\\
\cite{li2020incremental} & triangle & second-order & yes & linear motion\\
\cite{zhang2023provably} & convex & first-order & yes & articulated\\
\cite{hauser2021semi} & point & second-order & no & articulated\\
\cite{kalakrishnan2011stomp,zucker2013chomp} & point-penalty & first-order & no & articulated\\
\cite{ni2022robust} & convex & first-order & yes & articulated\\
Ours & convex & second-order & yes & articulated\\
\bottomrule
\end{tabular}
\caption{\label{table:features}\small{Comparison of planning methods across four dimensions: (Scalability) granularity (convex$>$triangle$>$point) of collision constraints, finer granularity can result in a larger number of constraints with limited scalability (certain algorithms use point-wise soft penalty and do not involve hard constraints, denoted as point-penalty); (Efficacy) type/convergence-speed of underlying numerical algorithm; (Robustness) whether collision constraints are guaranteed to be satisfied; (Generality) allowed type of robot motions.}
}
\end{table}
\TE{Main Results:} We present two innovative optimization solvers designed for collision-constrained robot pose and trajectory planning. Our approach highlights a structural analysis of the barrier function governing collision constraints between pairs of convex hulls. We demonstrate that the Hessian matrix of this barrier function possesses a unique structure that enables efficient factorization using a Schur-complement solver, leading to our ECB method. Furthermore, by employing a generalized barrier function, we eliminate the need for a separating plane between convex hull pairs by representing it as an implicit function of robot poses, leading to our ICB method. Both of these advancements empower us to apply second-order convergent Newton's methods with an iterative cost linear in the number of convex hull pairs. 

Our method exhibits several advantageous properties that have not been simultaneously achieved by prior algorithms. First, we decompose the robot's geometry into convex hulls rather than points~\cite{hauser2021semi} or triangles~\cite{li2020incremental}. As a result, our method can scale to large, complex robot and environmental geometries using a moderate number of geometric primitives, leading to enhanced scalability. Second, in contrast to previous first-order convergent methods~\cite{escande2014strictly,zhang2023provably}, our approach enjoys faster second-order convergence. Finally, our method adopts the interior point algorithm that inherits the robustness established by~\cite{li2020incremental,zhang2023provably}, i.e., our method ensures the collision-free properties of the optimize robot poses/trajectories. We have conducted evaluations of our method across six challenging robot planning scenarios. Our results unequivocally demonstrate that our approach achieves significantly faster convergence compared to the first-order baseline method.
\begin{figure*}[th]
\centering
\includegraphics[width=.99\linewidth]{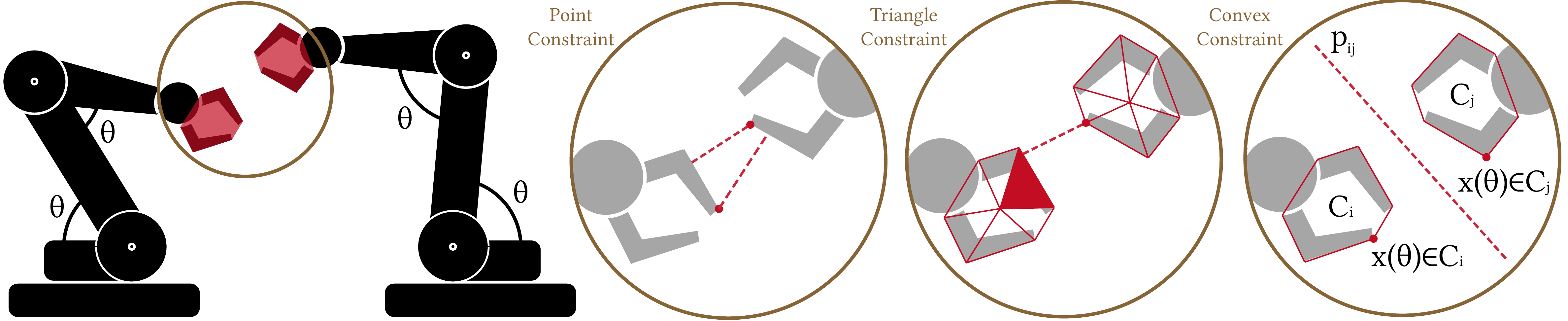}
\caption{\label{fig:pipeline}\small{We exemplify the optimization process involving two articulated robot poses, wherein the two end-effectors (red) must maintain a certain separation distance. Previous approaches~\cite{gao2017gradient,hauser2021semi} substitute such collision constraints with point constraints, while \cite{li2020incremental} suggests discretizing the end-effector using a triangle mesh and employing triangle constraints. In contrast, our method builds upon the convex constraints formulation introduced by \cite{ni2022robust,honig2018trajectory}, where a separating plane denoted as $p_{ij}$ is jointly optimized along with the robot poses. While~\cite{ni2022robust,honig2018trajectory} are first-order convergent methods, our approach stands out with its local second-order convergence.}}
\vspace{-10px}
\end{figure*}
\section{\label{eq:related}Related Work}
We review related work on collision-constrained planning, collision constraint formulations, numerical optimization techniques, and semi-infinite collision constraints.

\TE{Collision-Constrained Planning:} The algorithms summarized in~\prettyref{table:features} predominantly fall into the category of local optimization methods. These approaches leverage (high-order) derivatives of the objective/constraint functions to efficiently identify locally optimal plans. In contrast, optimal sampling-based methods~\cite{karaman2011sampling,janson2015fast} pursue globally optimal plans by exploring the entire configuration space. However, a notable drawback of these methods is their limited scalability concerning the dimensionality of the configuration space~\cite{janson2018deterministic}. An additional category of techniques includes the control barrier function~\cite{ames2019control}, designed for dynamic systems to adhere to specific constraints. However, these algorithms primarily focus on constraint satisfaction and do not directly address the planning of robot motions that fulfill users' requirements. We also acknowledge recent advancements~\cite{amice2022finding,marcucci2022motion} that involve precomputing feasible sub-domains within the configuration space. Subsequently, mixed-integer second-order numerical solvers are employed to derive globally optimal motion plans restricted to these sub-domains. This approach represents a promising alternative to our method, which obviates the need for sub-domain precomputation or restriction.

\TE{Collision-Constraint Formulations:} The local optimization techniques outlined in~\prettyref{table:features} can be distinguished by the granularity at which they formulate collision constraints. Several methods~\cite{gao2017gradient,hauser2021semi} opt to replace collision constraints with point constraints, due to their neat closed-form derivatives. However, this approach necessitates a considerable number of point constraints to ensure the safety of a robot with complex geometries. Recent advancements, as exemplified in~\cite{li2020incremental}, extend point constraints to collision constraints between pairs of triangles, benefiting from closed-form second-order derivatives. Nevertheless, even this approach requires a substantial number of triangles to represent intricate robot geometries, resulting in a high volume of collision constraints. As proposed in~\cite{honig2018trajectory,ni2022robust}, the number of geometric primitives can be significantly reduced if we can formulate collision constraints between general convex hulls. While \citewithauthor{escande2014strictly} demonstrated that the distance function between convex hulls possesses first-order derivatives, a comprehensive exploration of second-order differentiability and the efficient evaluation of convex constraints remains an area that requires further investigation.

\TE{Numerical Optimization Techniques:} Methods from~\prettyref{table:features} harness the power of numerical optimization solvers to search for optimal plans. The majority of existing approaches~\cite{gao2017gradient,deits2015efficient,liu2017planning} cast the problem in the form of a conventional (mixed-integer) Nonlinear Program (NLP) and subsequently employ off-the-shelf optimization software packages~\cite{gill2005snopt,biegler2009large}. These algorithms typically exhibit second-order convergence when an accurate Hessian matrix can be evaluated. However, they regress to first-order convergence when solely gradient information is accessible, as in~\cite{escande2014strictly}. Furthermore, customized optimization solvers have been devised to enhance robustness or efficiency. For instance, \citewithauthor{li2020incremental} seamlessly integrated continuous collision checks into the line-search procedure to circumvent the challenge of tunneling. Meanwhile, methods like~\cite{hauser2021semi,zhang2023provably} strategically sample collision constraints along the trajectory to approximate the solution of a Semi-Infinite Program (SIP). \citewithauthor{ni2022robust} introduced a first-order ADMM approach that interleaves the optimization of robot trajectories and the separating planes between convex hulls to facilitate parallel computation.

\TE{Semi-Infinite Collision Constraints:} As clarified in~\prettyref{sec:introduction}, the necessity to satisfy collision constraints at infinitely many time instances inherently gives rise to SIP problem instances, a concept recently articulated by~\cite{hauser2021semi,zhang2023provably}. While this paper primarily addresses a finite set of collision constraints, we emphasize that our methodology can be adapted to tackle SIP problems. Notably, existing SIP solvers propose to transform SIP into a standard NLP by either sampling~\cite{hauser2021semi} or discretizing~\cite{zhang2023provably} the constraints. In this context, we can readily substitute the underlying solver in~\cite{hauser2021semi,zhang2023provably} with our approach to achieve second-order convergence when dealing with convex constraints. However, it is noteworthy that, while this enables second-order convergence in the NLP subproblem, the convergence speed of the original SIP problem remains an open question and warrants further exploration, which we leave as a subject for future work.
\section{Problem Statement}
We address the challenge of collision-constrained optimization, focusing on the task of optimizing the pose/trajectory of a robot within the context of a twice-differentiable objective function $\mathcal{O}(\theta)$, where $\theta$ represents the robot's configuration parameter. The robot is subject to a set of collision constraints taking the following general form:
\begin{align*}
C_i(\theta)\cap C_j(\theta)=\emptyset
\quad\forall ij\in\mathcal{C},
\end{align*}
Here, $C_i$ and $C_j$ represent pairs of potentially contacting objects drawn from the set of contact pairs denoted as $\mathcal{C}$. $C_\bullet$ can correspond to a link within an articulated robot or a static obstacle. For the sake of generality, we assume that $\mathcal{O}(\theta)$ is a function of the robot's configuration. Additionally, we presume that $C_\bullet$ possesses a (not necessarily strictly) convex shape, which is a widely adopted geometric representation for robots~\cite{escande2014strictly,deits2015computing,amice2022finding}. Using convex shapes can effectively reduce the number of constraints as compared with point-pair or triangle-pair constraints. Non-convex shapes can be further decomposed into convex primitives. Consequently, the primary optimization model addressed in this paper can be summarized as:
\begin{align}
\label{eq:opt}
\argmin{\theta}\mathcal{O}(\theta)\quad\ST
C_i(\theta)\cap C_j(\theta)=\emptyset
\quad\forall ij\in\mathcal{C}.
\end{align}
Equation~\ref{eq:opt} embraces a generalized formulation that encompasses both the optimization of a static robot pose and a dynamic robot trajectory. In the case of optimizing a static robot pose, $\theta$ corresponds to a single configuration, and $\mathcal{C}$ is a finite set that enumerates all potential convex object pairs susceptible to collision. In contrast, when optimizing a robot trajectory, $\theta$ represents the parameterization of the robot's trajectory, such as the control points of a spline curve~\cite{deits2015efficient,zhou2019robust}. In this scenario, $\mathcal{C}$ expands to an infinite set that encompasses all possible convex object pairs at infinitely many time instances across the trajectory. This extension leads to the formulation of SIP problem instances. In the latter case, $\mathcal{C}$ can be further approached by sampling at finite time instances~\cite{hauser2021semi} or discretizing~\cite{zhang2023provably}, thus unified with the former case.

\section{Method}
The effective resolution of~\prettyref{eq:opt} poses practical challenges, particularly when dealing with convex collision constraints known for their non-differentiability~\cite{escande2014strictly,hauser2021semi}. As illustrated in~\prettyref{fig:pipeline}, when complex environmental geometries are considered, the number of collision pairs can rapidly escalate, leading to sizable problem dimensions. In this study, we introduce two efficient optimization techniques, both featuring local second-order convergence and an iterative cost linear in the number of collision pairs. Our approach draws inspiration from recent research~\cite{honig2018trajectory,ni2022robust}, which introduced separating planes between pairs of convex objects, represented by the normal direction $n_{ij}$ and offset $d_{ij}$. However, prior works~\cite{honig2018trajectory,ni2022robust} employed an Alternating Optimization (AO) approach that interleaved the updates of $n_{ij},d_{ij}$ and $\theta$. Unfortunately, this approach is limited to at most first-order convergence speed~\cite{xu2017globally}. Our primary contribution is the development of two quasi-Newton algorithms that jointly optimize $n_{ij},d_{ij}$, and $\theta$ without significantly increasing the computational overhead. To achieve this, we augment our objective function with a barrier energy designed to prevent the intersection of $C_i$ and $C_j$. In our ECB method, we derive the Hessian matrix of the augmented objective function concerning both $n_{ij},d_{ij}$, and $\theta$. Subsequently, we efficiently invert this large Hessian matrix using the Schur-complement lemma~\cite{boyd2004convex}. In our ICB method, we eliminate $n_{ij}$ and $d_{ij}$ by expressing them as implicit functions of $\theta$ and derive its Hessian using the implicit function theorem~\cite{agrawal2019differentiable}.

\subsection{ECB Method}
A convex object can be represented in two ways: the V-representation characterizes $C_\bullet$ using a set of vertices, while the H-representation characterizes it using a set of separating planes. In this work, we consistently utilize the V-representation. Without any ambiguity, we denote $C_\bullet$ as the set of vertices. We assume each vertex $x(\theta)\in C_\bullet$ is a twice-differentiable function of $\theta$, which holds generally for articulated robot with $x(\theta)$ being the forward-kinematic function. To define a valid separating plane $p_{ij}\triangleq\TWO{n_{ij}}{d_{ij}}$, it should ensure that the vertices of $C_i$ and $C_j$ are on opposite sides of it, i.e.:
\begin{align*}
n_{ij}x+d_{ij}<0\;\forall x\in C_i\land
n_{ij}x+d_{ij}>0\;\forall x\in C_j.
\end{align*}
We enforce each of these linear constraints by introducing a barrier energy function $P(\bullet)$ and augmenting the objective function as follows:
\small
\begin{align*}
\bar{\mathcal{O}}(\theta,\cdots,p_{ij},\cdots)=&\mathcal{O}(\theta)+\sum_{ij\in\mathcal{C}}P_{ij}(\theta,p_{ij})\\
P_{ij}(\theta,p_{ij})\triangleq&\sum_{x\in C_i}P(-n_{ij}x(\theta)-d_{ij})+\sum_{x\in C_j}P(n_{ij}x(\theta)+d_{ij}),
\end{align*}
\normalsize
where we assume the barrier function $P(\bullet)$ is strictly convex, defined in the interval $(0,\infty)$, and satisfies the condition $\lim_{\bullet\to0^+}P(\bullet)=\infty$. Consequently, our primary optimization problem~\prettyref{eq:opt} is transformed into the following barrier-augmented optimization:
\begin{align*}
\argmin{\theta,p_{\bullet}}\;\bar{\mathcal{O}}(\theta,\cdots,p_{ij},\cdots)\quad\ST\|n_{ij}\|=1\quad\forall ij\in\mathcal{C}.
\end{align*}
Note that we still need an additional constraint to ensure unit normal vectors for the separating planes. This method was previously proposed in~\cite{ni2022robust}. However, they optimized $p_{ij}$ and $\theta$ in separate sub-problems. In contrast, we introduce a joint optimization formulation for both $p_{ij}$ and $\theta$, achieving faster convergence. We refer to this method as the ECB method since all variables are explicitly treated as independent decision variables.

Our approach to optimizing the explicit formulation employs a standard Newton's method. We calculate the joint Hessian matrix $\nabla^2\bar{\mathcal{O}}$ and then determine the joint update direction $\delta\triangleq\FOUR{\delta\theta^T}{\cdots}{\delta p_{ij}^T}{\cdots}^T$ by solving the following KKT system:
\begin{align}
\label{eq:KKT}
0=\left(\setlength{\arraycolsep}{1pt}
\begin{array}{cc}
\nabla^2\bar{\mathcal{O}} & J\\
J^T & 0
\end{array}\right)
\TWOC{\delta}{\lambda}+\TWOC{\nabla\bar{\mathcal{O}}}{0},
\end{align}
where $J$ represents the Jacobian of the unit normal constraint $\|n_{ij}\|=1$, with each column of $J$ corresponding to the normal vector of a separating plane. However, this straightforward approach can encounter two main issues. Firstly, when dealing with complex environmental geometries, the number of separating planes can reach tens of thousands~\cite{zhang2023provably}, resulting in large Hessian matrices. If we use standard direct factorization methods to invert such matrices, the computational overhead can compromise the advantages of second-order convergence. Additionally, the Hessian matrix can exhibit exceedingly large condition numbers, leading to numerical instability. Fortunately, we can leverage the unique structure of the Hessian matrix to develop an efficient and stable matrix factorization scheme with linear complexity in the number of separating planes.

We first notice that a separating plane only appears in a single term $P_{ij}$. Therefore, the Hessian matrix takes the following form:
\begin{align*}
\nabla^2\bar{\mathcal{O}}=
\left(\setlength{\arraycolsep}{1pt}
\begin{array}{cccc}
\nabla_{\theta\theta}\bar{\mathcal{O}} & \cdots & \nabla_{\theta p}P_{ij} & \cdots\\
\vdots & \ddots & & \\
\nabla_{p\theta}P_{ij} & & \nabla_{pp}P_{ij} & \\
\vdots & & & \ddots
\end{array}\right).
\end{align*}
We propose a Gauss elimination approach to solve the KKT system in~\prettyref{eq:KKT} and show that this procedure is well-defined. We first notice that $\lambda$ can also be decomposed into $\lambda_{ij}$, each related to one separating plane. The joint linear sub-system of $p_{ij},\lambda_{ij}$ takes the following form:
\begin{align*}
\TWOC{\nabla_{p\theta}P_{ij}}{0}\delta\theta+
\left(\setlength{\arraycolsep}{1pt}
\begin{array}{cc}
\nabla_{pp}P_{ij} & n_{ij}\\
n_{ij}^T & 0
\end{array}\right)
\TWOC{\delta p_{ij}}{\lambda_{ij}}=
\TWOC{\nabla_pP_{ij}}{0},
\end{align*}
where we abuse notation and add a pending zero to $n_{ij}$ when necessary. Suppose the coefficient matrix is invertible, we can solve for $\delta p_{ij}$ as:
\begin{equation}
\begin{aligned}
\label{eq:schur1}
\delta p_{ij}=&H_{ij}\left(\nabla_pP_{ij}-\nabla_{p\theta}P_{ij}\delta\theta\right)\\
H_{ij}\triangleq&\nabla_{pp}P_{ij}^{-1}
\left[I-n_{ij}n_{ij}^T\frac{\nabla_{pp}P_{ij}^{-1}}{n_{ij}^T\nabla_{pp}P_{ij}^{-1}n_{ij}}\right],
\end{aligned}
\end{equation}
where we have used Gauss elimination to factor out $\lambda_{ij}$ assuming $\nabla_{pp}P_{ij}$ is invertible. Next, we plug all the $\delta p_{ij}$ into the first equation to yield the following Schur-complement system of $\theta$ alone:
\begin{align*}
H_\theta\delta\theta=&\left[\nabla_\theta\bar{\mathcal{O}}-\sum_{ij\in\mathcal{C}}
\nabla_{\theta p}P_{ij}H_{ij}\nabla_{p}P_{ij}\right]\numberthis\label{eq:schur2}\\
H_\theta\triangleq&\left[\nabla_{\theta\theta}\bar{\mathcal{O}}-\sum_{ij\in\mathcal{C}}
\nabla_{\theta p}P_{ij}H_{ij}\nabla_{p\theta}P_{ij}\right],
\end{align*}
which is then solved for $\theta$. Notably, the above procedure has a complexity linear in the number of hyper planes, which is much faster than using an off-the-shelf solver to factorize the large KKT matrix in~\prettyref{eq:KKT}. However, the well-definedness of the above system relies on the inversion of the matrix $\nabla_{pp}P_{ij}$ and $H_{\theta}$. Using the Schur-complement lemma~\cite{boyd2004convex}, we have the following corollary (proved in appendix):
\begin{corollary}
\label{cor:explicit}
If $\nabla^2\bar{\mathcal{O}}\succeq\epsilon I$, then the Schur-complementary solver is well-defined and $H_\theta\succeq\epsilon I$.
\end{corollary}
In practice, however, the strict requirement of $\nabla^2\bar{\mathcal{O}}\succeq\epsilon I$ may not be met, and we introduce a diagonal perturbation to ensure well-definedness. Specifically, we employ an adjustment function $\mathcal{A}(H,\epsilon)$ that ensures all eigenvalues of $H$ to be greater than a specified $\epsilon$. This adjustment is applied to both $\nabla_{pp}P_{ij}$ and $H_\theta$. The primary challenge in this adjustment process is the eigen decomposition. Thankfully, this operation is computationally efficient due to the relatively small matrix sizes involved. Following the computation of $\delta$, we utilize a line-search strategy to find a step size that strictly decreases the objective function. Finally, we re-normalize $n_{ij}$ after the update. The algorithm pipeline for the ECB method is outlined in~\prettyref{alg:explicit}.
\begin{algorithm}[ht]
\caption{\label{alg:explicit}ECB Method}
\begin{algorithmic}[1]
\Require{Initial $\theta,p_{ij},\mathcal{C},\epsilon_1,\epsilon_2,\epsilon$}
\Ensure{Locally optimal $\theta$}
\For{$\|\nabla\bar{\mathcal{O}}\|_\infty>\epsilon$}
\State Compute derivatives in \prettyref{eq:KKT}
\For{$ij\in\mathcal{C}$}
\State $\nabla_{pp}P_{ij}\gets\mathcal{A}(\nabla_{pp}P_{ij},\epsilon_1)$
\State Form complement system for $p_{ij}$ (\prettyref{eq:schur1})
\EndFor
\State Form complement system for $\theta$ (\prettyref{eq:schur2})
\State $H_\theta\gets\mathcal{A}(H_\theta,\epsilon_2)$ then solve for $\delta\theta$
\For{$ij\in\mathcal{C}$}
\State Plug $\delta\theta$ in \prettyref{eq:schur1} for $\delta p_{ij}$
\EndFor
\State Use line-search procedure to find step size $\alpha$
\State Update $\theta\gets\theta+\alpha\delta\theta$
\For{$ij\in\mathcal{C}$}
\State Update $p_{ij}\gets p_{ij}+\alpha\delta p_{ij}$, $n_{ij}\gets n_{ij}/\|n_{ij}\|$
\EndFor
\EndFor
\end{algorithmic}
\end{algorithm}
\subsection{ICB Method}
The primary computational challenge encountered in the ECB method is the inversion of the large Hessian matrix. However, many variables correspond to separating planes, which are intermediary variables unrelated to the robot's motion itself. Hence, it is advantageous to eliminate these variables, leaving only $\theta$ as the independent variable. In this section, we demonstrate that it is indeed possible to express $p_{ij}$ as a well-defined, sufficiently smooth function of $\theta$. For this reason, we refer to this method as the ICB method. To initiate this discussion, we observe that each plane $p_{ij}$ only appears in the collision penalty terms. Therefore, the optimal $p_{ij}$ should satisfy the following optimization problem:
\begin{align*}
p_{ij}\in\argmin{n_{ij},d_{ij}}P_{ij}(\theta,p_{ij})\quad\ST\|n_{ij}\|=1.
\end{align*}
However, the function defined above does not guarantee a well-defined $p_{ij}(\theta)$ because the solution may not be unique due to the additional non-convex unit-norm constraint. To obtain a well-defined function $p_{ij}(\theta)$, we turn to an equivalent formulation based on the Support Vector Machine (SVM)~\cite{chen2005tutorial}. This formulation replaces the non-convex constraint $\|n_{ij}\|=1$ with a relaxed convex constraint $\|n_{ij}\|\leq1$, giving the following alternative definition:
\begin{align*}
p_{ij}\in\argmin{n_{ij},d_{ij}}P_{ij}(\theta,p_{ij})\quad\ST\|n_{ij}\|\leq1.
\end{align*}
It is evident that the above optimization is convex. However, convexity alone is insufficient to express $p_{ij}$ as a function of $\theta$ because there may not be a unique minimizer. Even if a unique solution exists, the resulting implicit function $p_{ij}(\theta)$ may not be differentiable. As pointed out in~\cite{agrawal2019differentiable}, general convex optimization problems only have sub-gradients. Therefore, we introduce the following smoothed optimization amenable to differentiability, employing our log-barrier function $P(\bullet)$ to handle the additional convex constraint:
\begin{align}
\label{eq:strictConvexOpt}
p_{ij}(\theta)\triangleq\argmin{n_{ij},d_{ij}}P_{ij}(\theta,p_{ij})+P(1-\|n_{ij}\|).
\end{align}
We have the several essential properties for~\prettyref{eq:strictConvexOpt} as summarized below (proved in appendix):
\begin{corollary}
\label{cor:implicit}
Assuming $C_i$ and $C_j$ are closed set, $C_i$ or $C_j$ has non-zero volume, i.e. $|C_i|\neq0$ or $|C_j|\neq0$, and $P(\bullet)$ is third-order differentiable, we have the following properties: 1) \prettyref{eq:strictConvexOpt} has a solution iff $C_i(\theta)\cap C_j(\theta)=\emptyset$; 2) $p_{ij}$ is a well-defined function of $\theta$; 3) $p_{ij}(\theta)$ is twice-differentiable; 4) $\lim_{\dist(C_i,C_j)\to0^+}P_{ij}(\theta,p_{ij}(\theta))=\infty$, with $\dist(\bullet,\bullet)$ being the distance between convex hulls.
\end{corollary}

\begin{algorithm}[ht]
\caption{\label{alg:implicit}ICB Method}
\begin{algorithmic}[1]
\Require{Initial $\theta,p_{ij},\mathcal{C},\epsilon_1,\epsilon$}
\Ensure{Locally optimal $\theta$}
\For{$\|\nabla\tilde{\mathcal{O}}\|_\infty>\epsilon$}
\For{$ij\in\mathcal{C}$}
\State $p_{ij}(\theta)\gets$solution to \prettyref{eq:strictConvexOpt}
\State Compute derivatives of $p_{ij}(\theta)$ via~\prettyref{eq:pijDeriv}
\EndFor
\State Form $\nabla\tilde{\mathcal{O}}$ and $\nabla^2\tilde{\mathcal{O}}$
\State Adjust $\nabla^2\tilde{\mathcal{O}}\gets\mathcal{A}(\nabla^2\tilde{\mathcal{O}},\epsilon_1)$
\State Form search direction $\delta\theta\gets-\nabla^{-2}\tilde{\mathcal{O}}\nabla\tilde{\mathcal{O}}$
\State Use line-search procedure to fine step size $\alpha$
\State Update $\theta\gets\theta+\alpha\delta\theta$
\EndFor
\end{algorithmic}
\end{algorithm}
\prettyref{cor:implicit} ensures the differentiable structure of $p_{ij}(\theta)$ as long as one of the convex sets is non-degenerate. This limitation is not significant for robotic applications because robot links practically always have non-zero volume. With this structure, we can define an implicit optimization approach and solve the following optimization problem:
\begin{equation}
\begin{aligned}
\label{eq:optI}
\argmin{\theta}\tilde{\mathcal{O}}(\theta)\triangleq\bar{\mathcal{O}}(\theta,\cdots,p_{ij}(\theta),\cdots),
\end{aligned}
\end{equation}
replacing the explicit variable $p_{ij}$ with the implicit function $p_{ij}(\theta)$. \prettyref{cor:implicit} 3) ensures that~\prettyref{eq:optI} has the twice-differentiable objective function and \prettyref{cor:implicit} 4) ensures that the feasibility of~\prettyref{eq:optI} implies the collision-free property. To evaluate such objective function at given $\theta$, we need to solve~\prettyref{eq:strictConvexOpt} for every pair of convex sets in close proximity. After the solution, we can evaluate the first- and second-order derivative of $p_{ij}(\theta)$ and $\tilde{\mathcal{O}}$ (appendix). The computational complexity of these equations is linear with respect to the number of separating planes, similar to our explicit algorithm. However, solving \prettyref{eq:strictConvexOpt} for each pair of convex sets may initially appear computationally burdensome. Fortunately, we can expedite the process by employing a warm-start strategy, where we retain the solution from the previous iteration. If this warm-start approach fails (for instance, if the solution from the last iteration is infeasible), we can establish a feasible initial estimate by computing the closest pair of points on $C_i$ and $C_j$. We then use this information to determine the initial separating plane, setting it as the middle sectioning plane. In practice, we can solve the subproblems in~\prettyref{eq:strictConvexOpt} with just a few Newton iterations, and all of these subproblems can be solved in parallel. The algorithmic workflow of the ICB method is summarized in~\prettyref{alg:implicit}. Since the Hessian matrix of $\tilde{\mathcal{O}}$ remains relatively small, we can calculate its dense eigen-decomposition, which is essential for matrix inversion and performing the positive-definite adjustment. This adjustment ensures that the minimal eigenvalue of the Hessian matrix is at least $\epsilon_1$.
\section{\label{sec:results}Evaluation}
We conducted our experiments using C++ on a single desktop machine equipped with an 8-core AMD EPYC 7K62 CPU. All available cores were utilized to facilitate parallel computations related to separating planes, such as the computation of per-plane Schur-complements in the ECB method and the evaluation of implicit functions and their derivatives in the ICB method. We also incorporated the widely used acceleration technique of broadphase collision checks, similar to the one employed in prior works~\cite{li2020incremental,ni2022robust}. This technique excludes distant pairs of convex hulls from consideration, introducing barrier terms only when the convex hulls are in close proximity. It is important to note that once inserted, these barrier terms are not removed, even if the pair of convex hulls later move apart. Throughout our analysis, we require our penalty function to be positive and strictly convex. One function satisfying our needs is the layered penalty function~\cite{harmon2009asynchronous}, which is used in our implementation. Finally, we use hyper-parameters $\epsilon_1=0.001$ and $\epsilon_2=0.001$ throughout our experiements.

Additionally, we integrated our solver with an implementation of the SIP solver from~\cite{zhang2023provably}. This allows us to solve SIP problem instances, ensuring collision-free trajectories at every continuous time interval, by leveraging their constraint discretization method. For comparison, both the ECB and ICB methods are compared with the baseline Alternating Optimization (AO) technique~\cite{ni2022robust}, which optimizes $\theta$ and $p_{ij}$ in separate subproblems, where we also optimize $p_{ij}$ subproblems in parallel. This is state-of-the-art optimization technique, allowing the optimization of arbitrary articulated robot poses under convex constraints, with the only difference being the convergence speed.

\begin{figure*}[th]
\centering
\scalebox{.85}{
\setlength{\tabcolsep}{1px}
\def\arraystretch{0}
\begin{tabular}{ccc}
\includegraphics[width=.32\linewidth]{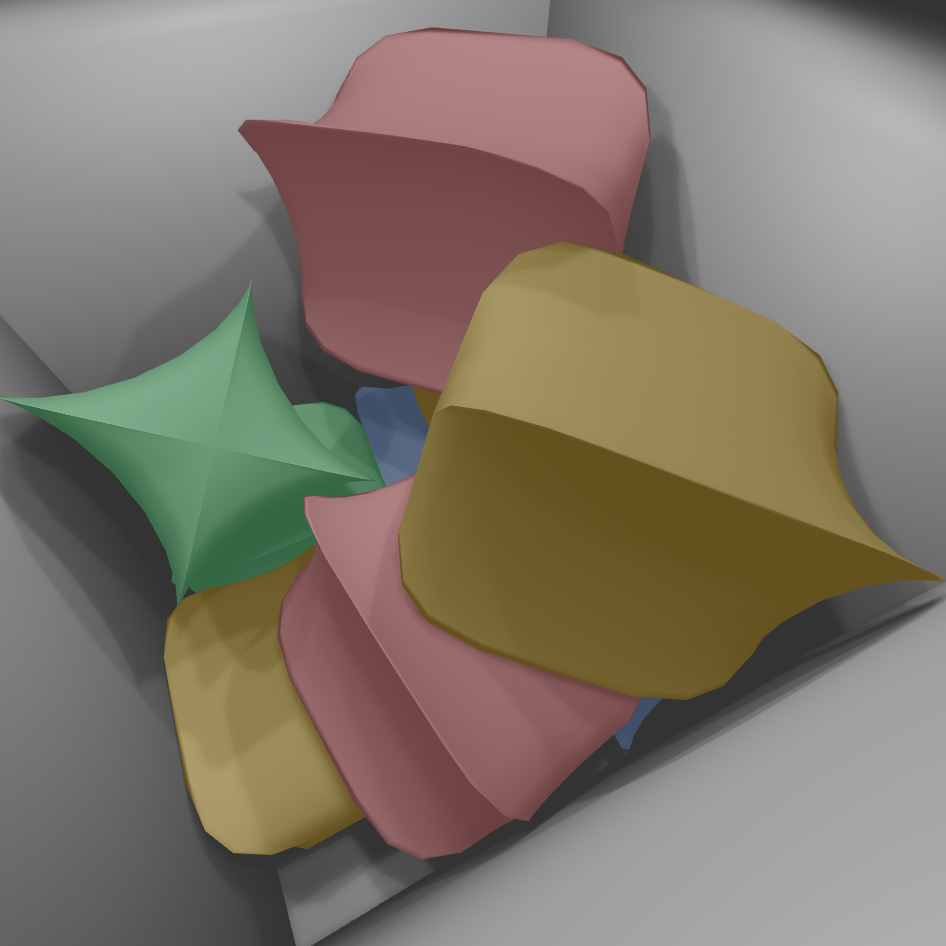}\put(-20,10){(a)} &
\includegraphics[width=.32\linewidth]{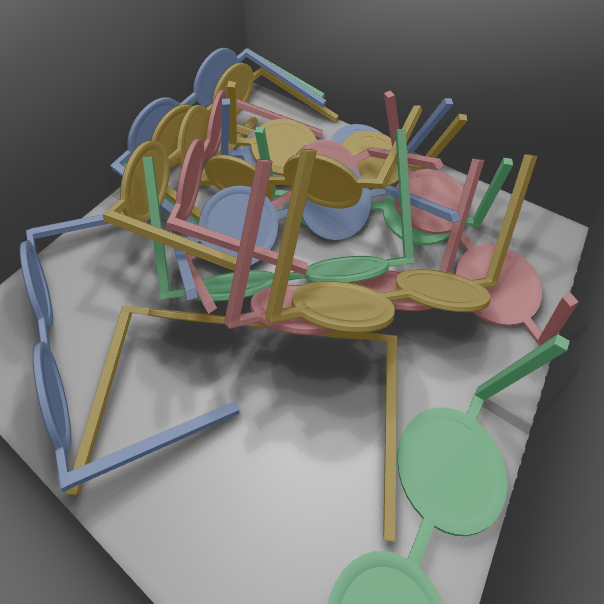}\put(-20,10){(b)} &
\includegraphics[width=.32\linewidth]{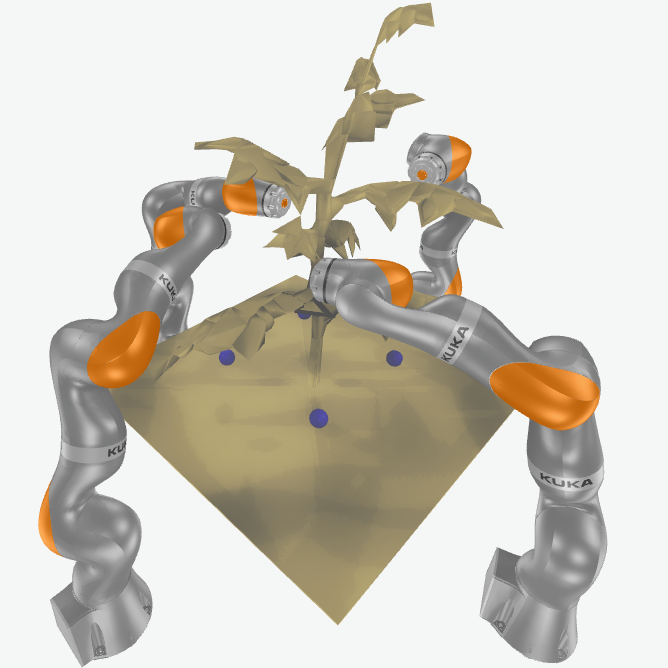}\put(-20,10){(c)}\\
\includegraphics[width=.3 \linewidth]{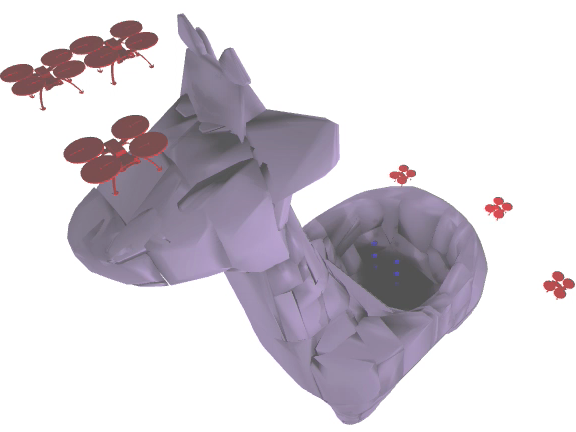}\put(-20,10){(d)} &
\includegraphics[width=.32\linewidth]{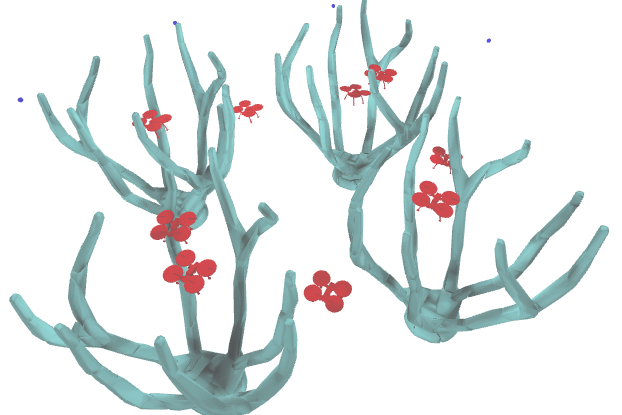}\put(-20,10){(e)} &
\includegraphics[width=.32\linewidth]{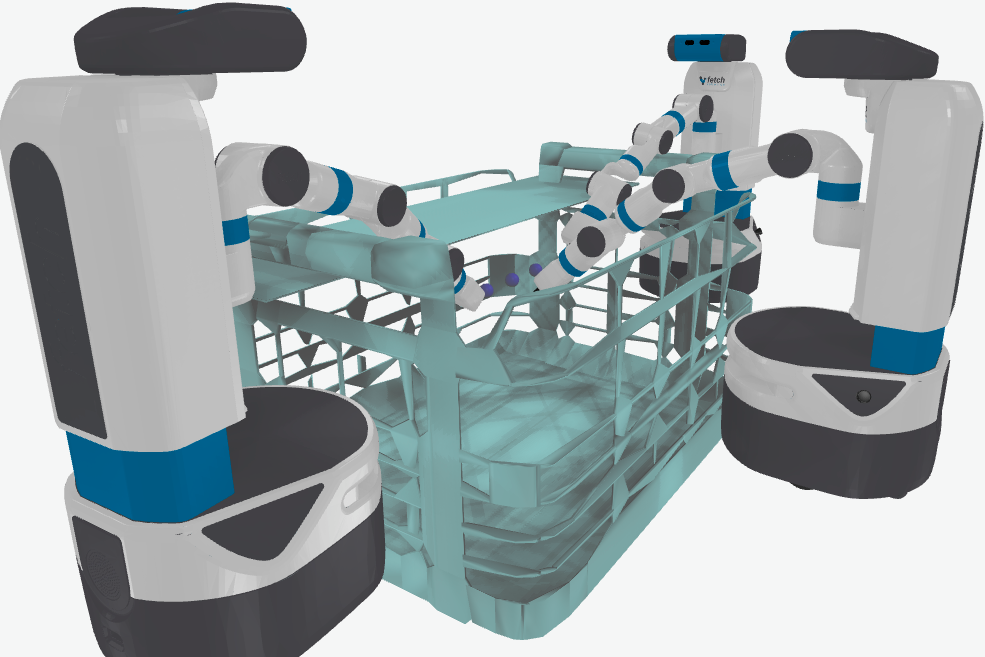}\put(-20,10){(f)}\\
\end{tabular}}
\caption{\label{fig:scenario}\small{The six scenarios used for evaluating the three methods: bulky object settling (a), thin object settling (b), quadruple-arm reaching (c), UAV trajectory generation (d), long-distance UAV trajectory generation (e), basket-reaching (f).}}
\end{figure*}
\begin{table}[ht]
\centering

\scalebox{0.95}
{
\begin{tabular}{c|c|ccc}
\toprule
 & $|\theta|$ & \multicolumn{3}{c}{$|\mathcal{C}|$}  \\
    &  & ECB & ICB & AO \\
\midrule
(a) &  54 &   50 &    49 &   50 \\
(b) & 108 &  716 &   570 &  714 \\
(c) &  24 & 1757 &  1024 & 1266 \\
(d) &  36 & 3099 &  3585 & 2444 \\
(e) &  54 & 6424 & 11122 & 6892 \\
(f) &  33 &  973 &   837 &  899 \\
\bottomrule
\end{tabular}
}
\caption{\label{table:complexity}\small{The number of degrees of freedom $|\theta|$ and separating planes $|\mathcal{C}|$ for each benchmarking scenario.} 
}
\vspace{-5px}
\end{table}

\begin{table*}[]
\centering
\scalebox{0.95}
{
\begin{tabular}{c|cccc|cccc|cccc}
\toprule
    & \multicolumn{4}{c|}{ECB} & \multicolumn{4}{c|}{ICB} & \multicolumn{4}{c}{AO} \\
Scenario & $\epsilon=10^{-1}$ & $\epsilon=10^{-2}$ & $\epsilon=10^{-3}$ & $\epsilon=10^{-4}$ & $\epsilon=10^{-1}$ & $\epsilon=10^{-2}$ & $\epsilon=10^{-3}$ & $\epsilon=10^{-4}$ &  $\epsilon=10^{-1}$ & $\epsilon=10^{-2}$ & $\epsilon=10^{-3}$ & $\epsilon=10^{-4}$ \\
\midrule
(a) & 697  &  703 & 5816 & 26906 &  337 &  337 & 1823 & 21610 & 24182 & 48386 &  N.A. &  N.A. \\
(b) & 6898 & 9814 & 9906 & 10329 & 1470 & 2063 & 4946 & 13245 & 42000 & 71561 & 77179 & 79300 \\
(c) & 424  & 1128 & 1474 &  2356 &  132 &  305 &  368 &   525 &   371 &   935 &  2046 &  N.A. \\
(d) & 1068 & 1479 & 1928 &  1987 &  406 &  528 &  576 &   623 &  1983 &  2941 &  4350 &  5010 \\
(e) & 1802 & 2383 & 2721 &  2936 & 1115 & 1277 & 1315 &  1396 &  2305 &  4808 &  5813 &  6773 \\
(f) & 219  &  435 &  604 &  1066 &  159 &  206 &  266 &   400 &   277 &  1896 &  N.A. &  N.A. \\
\bottomrule
\end{tabular}
}
\caption{\label{table:convIter}\small{The number of iterations for the three algorithms to reach difference level of gradient norm $\epsilon$ (N.A. denotes the algorithm cannot finish computation within the time limit).}}
\vspace{-5px}
\end{table*}


\begin{table*}[ht]
\centering
\scalebox{0.95}
{
\begin{tabular}{c|cccc|cccc|cccc}
\toprule
    & \multicolumn{4}{c|}{ECB} & \multicolumn{4}{c|}{ICB} & \multicolumn{4}{c}{AO} \\
Scenario & $\epsilon=10^{-1}$ & $\epsilon=10^{-2}$ & $\epsilon=10^{-3}$ & $\epsilon=10^{-4}$ & $\epsilon=10^{-1}$ & $\epsilon=10^{-2}$ & $\epsilon=10^{-3}$ & $\epsilon=10^{-4}$ &  $\epsilon=10^{-1}$ & $\epsilon=10^{-2}$ & $\epsilon=10^{-3}$ & $\epsilon=10^{-4}$ \\
\midrule
(a) &   0.04 &   0.04 &   0.36 &   1.70 &  0.02 &  0.02 &   0.10 &   1.44 &   1.14 &   2.32 &   N.A. &   N.A. \\
(b) &   3.58 &   5.15 &   5.20 &   5.40 &  0.74 &  1.07 &   2.73 &   7.54 &  20.63 &  34.88 &  37.59 &  38.56 \\
(c) &   2.58 &  11.20 &  17.29 &  38.05 &  0.76 &  2.37 &   3.17 &   5.51 &   1.78 &   6.19 &  23.09 &   N.A. \\
(d) &  13.44 &  20.75 &  29.61 &  30.80 &  4.99 &  7.47 &   8.52 &   9.56 &  27.28 &  43.90 &  68.96 &  81.06 \\
(e) & 107.88 & 164.47 & 200.00 & 222.79 & 76.74 & 95.76 & 100.23 & 109.99 & 161.08 & 446.63 & 565.25 & 679.28 \\
(f) &   2.57 &   6.56 &  10.96 &  26.73 &  1.72 &  2.58 &   3.96 &   7.71 &   4.47 &  63.96 &   N.A. &   N.A. \\
\bottomrule
\end{tabular}
}
\caption{\label{table:convTime}\small{The computational time (min) for the three algorithms to reach difference level of gradient norm $\epsilon$.}}
\vspace{-5px}
\end{table*}
We assess the performance of our method using a series of six challenging benchmarks, as outlined in~\prettyref{fig:scenario}. The first two benchmarks involve object settling problems, with bulky objects (a) and thin objects (b) dropped into a box. These problems require us to compute the static equilibrium poses of the objects. Since we optimize a single pose, these scenarios only involve a finite set of collision constraints. The remaining four scenarios focus on trajectory optimization problems, where we solve SIP problem instances by integrating both our methods and AO~\cite{ni2022robust} into the SIP solver framework~\cite{zhang2023provably}. In~\prettyref{fig:scenario} (c), four robot arms reach various end-effector positions on a table, while avoiding obstacles. In~\prettyref{fig:scenario} (d), we optimize the trajectories of six UAVs with a giraffe-shaped obstacle in the middle. \prettyref{fig:scenario} (e) extends this challenge with longer trajectories for the nine UAVs and multiple obstacles along their paths. Finally, \prettyref{fig:scenario} (f) presents three fetch robots reaching 3 positions in a basket. In these scenarios, the trajectories for each robot's configuration space are parameterized using high-order spline curves, with the number of degrees of freedom and separating planes summarized in~\prettyref{table:complexity}, which shows that our method can scale to complex high-DOF robots and environmental setups. We measure the number of iterations and computational costs required to reach different gradient norm values, and the results are summarized in~\prettyref{table:convIter} and~\prettyref{table:convTime}, respectively.

Our results clearly demonstrate the superior convergence speed of our method compared to AO, especially when aiming for highly optimal motion plans with a small threshold value ($\epsilon$). The key factor behind this superior performance is the significantly lower number of iterations required by our method, while maintaining a comparable per-iteration cost. Furthermore, it is worth noting that ICB exhibits faster convergence compared to ECB, despite both methods being locally second-order. This can be attributed to ICB's approach of computing the optimal separating planes during each iteration. Although this leads to a slightly higher iterative cost, the overall reduction in the number of iterations more than compensates for this additional computational overhead. In summary, our method consistently achieves a remarkable reduction in computational time, with a speedup ranging from $5.12$ to $206.34$ for pose optimization (\prettyref{fig:scenario}a-b) and $2.63$ to $151.23$ for trajectory optimization (\prettyref{fig:scenario}c-f).
\section{\label{sec:conclusion}Conclusion \& Limitations}
We present innovative second-order optimization-based planners for both robot poses and trajectories, building upon the convex constraint formulation introduced by~\cite{honig2018trajectory,ni2022robust}. A key distinguishing feature of our approach is the efficient application of Newton's method to simultaneously optimize the separating planes and robot poses. 
Our work has paved the way for several promising avenues in future research. Although both of our methods outperform the AO baseline, they each come with their own advantages and disadvantages. On the one hand, ICB is more efficient than ECB, requiring even fewer iterations to converge. However, it demands a more complex implementation with an intricate inner loop of optimization and a careful treatment of second-order derivatives. The reliable evaluation of these derivatives under finite-precision floating-point formats requires further investigation. Further, due to the inherent complexity of the underlying problem, our method may still be time-consuming.
\AtNextBibliography{\footnotesize}
\printbibliography
\section*{Appendix: Additional Proofs}
\begin{proof}[Proof of~\prettyref{cor:explicit}]
$\nabla^2\bar{\mathcal{O}}$ is positive definite and so is all its principle submatrices. As a result, $\nabla_{pp}P_{ij}$ is invertible and $n_{ij}^T\nabla_{pp}P_{ij}^{-1}n_{ij}\neq0$ so $H_{ij}$ is well-defined. We further rewrite:
\begin{align*}
H_{ij}=&\nabla_{pp}P_{ij}^{-\frac{1}{2}}
\left[I-\bar{n}_{ij}\bar{n}_{ij}^T\right]\nabla_{pp}P_{ij}^{-\frac{1}{2}}\\
\bar{n}_{ij}\triangleq&{\nabla_{pp}P_{ij}^{-\frac{1}{2}}n_{ij}}/{\|\nabla_{pp}P_{ij}^{-\frac{1}{2}}n_{ij}\|},
\end{align*}
from which we immediately have $\nabla_{pp}P_{ij}^{-1}\succeq H_{ij}\succeq0$. By Guttman rank additivity~\cite{CARLSON1986257}, we have the following matrix is positive definite:
\begin{align*}
\bar{H}_\theta\triangleq\left[\nabla_{\theta\theta}\bar{\mathcal{O}}-\sum_{ij\in\mathcal{C}}
\nabla_{\theta p}P_{ij}\nabla_{pp}P_{ij}^{-1}\nabla_{p\theta}P_{ij}\right].
\end{align*}
Combined with the fact that $\nabla_{pp}P_{ij}^{-1}\succeq H_{ij}$, we have:
$H_\theta\succeq\bar{H}_\theta\succ0$. Note the above argument only requires $\nabla^2\bar{\mathcal{O}}\succeq0$. If we further have: $\nabla^2\bar{\mathcal{O}}\succeq\epsilon I$, then we can apply the above argument to the perturbed matrix:
\begin{align*}
\left(\setlength{\arraycolsep}{1pt}
\begin{array}{cc}
\nabla^2\bar{\mathcal{O}} & J\\
J^T & 0
\end{array}\right)-\MTT{\epsilon I}{}{}{0},
\end{align*}
which immediately yields: $H_\theta\succeq\bar{H}_\theta\succeq\epsilon I$.
\end{proof}
\begin{proof}[Proof of~\prettyref{cor:implicit}]
1) If \prettyref{eq:strictConvexOpt} is feasible, then $n_{ij}\neq0$ because otherwise we have $P(-d_{ij})+P(d_{ij})<\infty$. But this implies $d_{ij}\in(0,\infty)$ and $-d_{ij}\in(0,\infty)$ which is impossible. Therefore, we can verify that $n_{ij}/\|n_{ij}\|,d_{ij}/\|n_{ij}\|$ is a plane separating $C_i$ and $C_j$. Conversely, if $C_i\cap C_j=\emptyset$, then $\dist(C_i,C_j)>0$, i.e., we have a separating plane $n_{ij}$ such that $-n_{ij}x(\theta)-d_{ij}>0$ for all $x(\theta)\in C_i$ and $n_{ij}x(\theta)+d_{ij}>0$ for all $x(\theta)\in C_j$. We can then choose $\alpha\in(0,1)$ sufficiently close to $1$ such that: $-\alpha n_{ij}x(\theta)-d_{ij}>0$ for $x(\theta)\in C_i$ and $\alpha n_{ij}x(\theta)+d_{ij}>0$ for $x(\theta)\in C_j$. And we can immediately verify that $p_{ij}=\TWOR{\alpha n_{ij}}{d_{ij}}$ is a feasible solution of~\prettyref{eq:strictConvexOpt}. 

2) This property essentially requires the uniqueness of minimizer. To establish uniqueness, we can write the Hessian of the objective function of~\prettyref{eq:strictConvexOpt} as:
\footnotesize
\begin{align*}
&\tilde{H}_{ij}(\theta,p_{ij})\triangleq\nabla_{p_{ij}}^2\left[P_{ij}(\theta,p_{ij})+P(1-\|n_{ij}\|)\right]\\
=&\sum_{x\in C_i}P''(-n_{ij}x(\theta)-d_{ij})\TWOC{x(\theta)}{1}\TWO{x(\theta)^T}{1}+\\
&\sum_{x\in C_j}P''(n_{ij}x(\theta)+d_{ij})\TWOC{x(\theta)}{1}\TWO{x(\theta)^T}{1}+
\nabla_{p_{ij}}^2\left[P(1-\|n_{ij}\|)\right].
\end{align*}
\normalsize
We show that $\tilde{H}_{ij}$ must be a matrix of full-rank. Note that $P''(-n_{ij}x(\theta)-d_{ij})>0$ and $P''(n_{ij}x(\theta)+d_{ij})>0$ due to the strong convexity of $P$. Next, note that $P(1-\|n_{ij}\|)$ is a convex function because it is a composite of a convex function $-1+\|n_{ij}\|$ and a non-decreasing function $P(-\bullet)$~\cite{boyd2004convex}. Further, we have the top-left entry of Hessian taking the following form:
\begin{align*}
\FPPT{P(1-\|n_{ij}\|)}{n_{ij,x}}=\frac{n_{ij,y}^2+n_{ij,z}^2}{\|n_{ij}\|^3(1-\|n_{ij}\|)}+\frac{n_{ij,x}^2}{\|n_{ij}^2\|(1-\|n_{ij}\|)}>0.
\end{align*}
This implies the Hessian of $P(1-\|n_{ij}\|)$ is a positive semi-definite, non-zero matrix, so it has the following diagonalization:
\begin{align*}
\FPPT{P(1-\|n_{ij}\|)}{n_{ij}}=V\Sigma V^T,
\end{align*}
where $0\neq\Sigma\succeq0$ and $V$ is a orthogonal matrix. Putting all these facts together, we have:
\begin{align*}
\tilde{H}_{ij}=&UU^T\\
U\triangleq&\THREE{U_1}{U_2}{U_3}\\
U_1\triangleq&\THREE{\cdots}{\sqrt{P''(-n_{ij}x(\theta)-d_{ij})}\TWOC{x(\theta)\in C_i}{1}}{\cdots}\\
U_2\triangleq&\THREE{\cdots}{\sqrt{P''(n_{ij}x(\theta)+d_{ij})}\TWOC{x(\theta)\in C_j}{1}}{\cdots}\\
U_3\triangleq&\left(\setlength{\arraycolsep}{1pt}\begin{array}{ccc}
V_1 & V_2 & V_3\\
0 & 0 & 0\\
\end{array}\right)\sqrt{\Sigma}.
\end{align*}
Without a loss of generality, we can assume $|C_i|\neq0$ and $\Sigma_{11}>0$. Then $x(\theta)\in C_i$ spans $\mathbb{R}^3$ and hence the following vectors span $\mathbb{R}^4$:
\begin{align*}
\FOUR{\cdots}{\sqrt{P''(-n_{ij}x(\theta)-d_{ij})}\TWOC{x(\theta)\in C_i}{1}}{\cdots}{\TWOC{V_1\sqrt{\Sigma_{11}}}{0}}.
\end{align*}
Put together, we have $U$ spans $\mathbb{R}^4$ and $\tilde{H}_{ij}=UU^T\succ0$. Due to the arbitarity of $p_{ij}$, we have~\prettyref{eq:strictConvexOpt} is a strictly convex optimization and $p_{ij}$ is a unique minimizer. 

3) Due to the strict convexity, $p_{ij}(\theta)$ is equivalently defined as the solution of:
\begin{align*}
\tilde{G}_{ij}(\theta,p_{ij})\triangleq\nabla_{p_{ij}}\left[P_{ij}(\theta,p_{ij})+P(1-\|n_{ij}\|)\right]=0.
\end{align*}
Since $\tilde{H}_{ij}(\theta,p_{ij})$ has full-rank, we can then invoke the high-order implicit function theorem to see that $p_{ij}(\theta)$ is a twice-differentiable function of $\theta$ if the barrier $P(\bullet)$ is third-order differentiable. 

4) Denote $d\triangleq\dist(C_i,C_j)>0$, then for every separating plane $p_{ij}$ with $\|n_{ij}\|=1$, we have $-n_{ij}x(\theta)-d_{ij}<d$ for some $x(\theta)\in C_i$ and $n_{ij}x(\theta)+d_{ij}<d$ for some $x(\theta)\in C_j$ because otherwise $\dist(C_i,C_j)>d$. Now the solution to~\prettyref{eq:strictConvexOpt} can be denoted as a generalized separating plane with $\|n_{ij}\|<1$, so we have $-n_{ij}x(\theta)-d_{ij}<d\|n_{ij}\|<d$ for some $x(\theta)\in C_i$ and $n_{ij}x(\theta)+d_{ij}<d\|n_{ij}\|<d$ for some $x(\theta)\in C_j$. As a result we have: $\lim_{d\to0^+}P_{ij}(\theta,p_{ij}(\theta))\geq\lim_{d\to0^+}P(d)=\infty$.
\end{proof}
\section*{Appendix: Derivatives}
We use Einstein's notation with scripts $\alpha,\beta,\gamma$:
\small
\begin{equation}
\begin{aligned}
\label{eq:pijDeriv}
&\FPP{p_{ij}(\theta)}{\theta^\alpha}=-\tilde{H}_{ij}^{-1}\FPP{\tilde{G}_{ij}}{\theta^\alpha}\\
&\FPPTT{p_{ij}(\theta)}{\theta^\alpha}{\theta^\beta}=-\tilde{H}_{ij}^{-1}\Big[
\FPPTT{\tilde{G}_{ij}}{\theta^\alpha}{\theta^\beta}+\FPP{\tilde{H}_{ij}}{\theta^\alpha}\FPP{p_{ij}(\theta)}{\theta^\beta}+\\
&\FPP{\tilde{H}_{ij}}{\theta^\beta}\FPP{p_{ij}(\theta)}{\theta^\alpha}+\FPP{\tilde{H}_{ij}}{p_{ij}^\gamma}\FPP{p_{ij}(\theta)}{\theta^\alpha}\FPP{p_{ij}^\gamma(\theta)}{\theta^\beta}\Big]\\
&\FDD{\tilde{\mathcal{O}}}{\theta}=\FPP{\tilde{\mathcal{O}}}{\theta}+\sum_{ij\in\mathcal{C}}\FPP{\tilde{\mathcal{O}}}{p_{ij}}\FPP{p_{ij}(\theta)}{\theta}
\quad\FDDT{\tilde{\mathcal{O}}}{\theta}=\FPPT{\tilde{\mathcal{O}}}{\theta}+\\
&\sum_{ij\in\mathcal{C}}\Big[\FPPTT{\tilde{\mathcal{O}}}{\theta}{p_{ij}}\FPP{p_{ij}(\theta)}{\theta}+
\FPP{p_{ij}(\theta)}{\theta}^T\FPPTT{\tilde{\mathcal{O}}}{p_{ij}}{\theta}+\\
&\FPP{p_{ij}(\theta)}{\theta}^T\FPPT{\tilde{\mathcal{O}}}{p_{ij}}\FPP{p_{ij}(\theta)}{\theta}+
\FPP{\tilde{\mathcal{O}}}{p_{ij}^\alpha}\FPPT{p_{ij}^\alpha(\theta)}{\theta}\Big].
\end{aligned}
\end{equation}
\end{document}